\pdfoutput=1

\documentclass[11pt]{article}

\usepackage[final]{acl}

\usepackage{natbib}
\usepackage{times}
\usepackage{latexsym}
\usepackage{amsmath}
\usepackage{multirow}
\usepackage{svg}
\usepackage{caption}
\usepackage{subcaption}
\usepackage{booktabs}
\usepackage{xcolor}
\usepackage[T1]{fontenc}

\usepackage[utf8]{inputenc}

\usepackage{microtype}

\usepackage{inconsolata}
\usepackage{enumitem}
\usepackage{graphicx}
\usepackage{amssymb}
\usepackage{bbm}

\usepackage{color, colortbl}
\definecolor{my_gray}{gray}{0.9}
\definecolor{gray}{rgb}{0.5,0.5,0.5}

\newcommand{\att}{$A^3T$} 
\newcommand{\illama}{IL-LLaMA3}

\newcommand\torevise[1]{\textcolor{blue}{#1}}
\newcommand{\modelname}{Retrospex} 

\title{Retrospex: Language Agent Meets Offline Reinforcement Learning Critic}
\vspace{5cm}
\author{Yufei Xiang \qquad  Yiqun Shen \qquad Yeqin Zhang\qquad  Cam-Tu Nguyen\\
         State Key Laboratory for Novel Software Technology, Nanjing University\\ School of Artificial Intelligence, Nanjing University\\ Nanjing, China \\ \texttt{\{xiangyf, yiqunshen, zhangyeqin\}@smail.nju.edu.cn}\\
         \texttt{ncamtu@nju.edu.cn}}

\begin{document}
\maketitle
\begin{abstract}

Large Language Models (LLMs) possess extensive knowledge and commonsense reasoning capabilities, making them valuable for creating powerful agents. However, existing LLM agent frameworks have not fully utilized past experiences for improvement. This work introduces a new LLM-based agent framework called \modelname{} , which addresses this challenge by analyzing past experiences in depth. Unlike previous approaches, \modelname{} does not directly integrate experiences into the LLM's context. Instead, it combines the LLM's action likelihood with action values estimated by a Reinforcement Learning (RL) Critic, which is trained on past experiences through an offline ``retrospection'' process. Additionally, \modelname{}  employs a dynamic action rescoring mechanism that increases the importance of experience-based values for tasks that require more interaction with the environment. We evaluate \modelname{}  in ScienceWorld, ALFWorld and Webshop environments, demonstrating its advantages over strong, contemporary baselines\footnote{https://github.com/Yufei-Xiang/Retrospex}.

\end{abstract}
\section{Introduction}


The emergence of LLMs has paved the way for the development of LLM-based agents. These agents leverage the vast knowledge and commonsense reasoning capabilities within LLMs to tackle a wide range of tasks \cite{ wang2022scienceworld, yao2022webshop,shridhar2020alfworld,yang2018hotpotqa,thorne2018fever}. Despite their potential, a significant challenge arises from the dependence on general-purpose LLMs. Specifically, these agents might not be sufficiently adapted to the specific environments, potentially hindering their task completion effectiveness.


Training LLM-based agents for new environments poses significant challenges. A common approach is to fine-tune the LLM using ``correct'' demonstrations—sample trajectories that successfully complete the task \cite{qin2024toolllm, lin2024swiftsage, zeng2023agenttuning}.
However, this approach focuses on correct behaviors, limiting the agent’s ability to learn and recover from mistakes.
Recently, efforts have been made to leverage imperfect experiences for training LLM agents. These methods fall into two categories: those that rely on working memory, like Reflexion \cite{shinn2023reflexion}, and those that utilize cross-task experiences from long-term memory, such as Rememberer \cite{zhang2024large}. Despite the progress, experiences are still not sufficiently used, as they are only integrated into the LLM’s context. Due to the limited context length of LLMs, this constrains the inclusion of more comprehensive experiences.


\begin{figure*}[]
    \centering
    \includegraphics[width=0.99\textwidth]{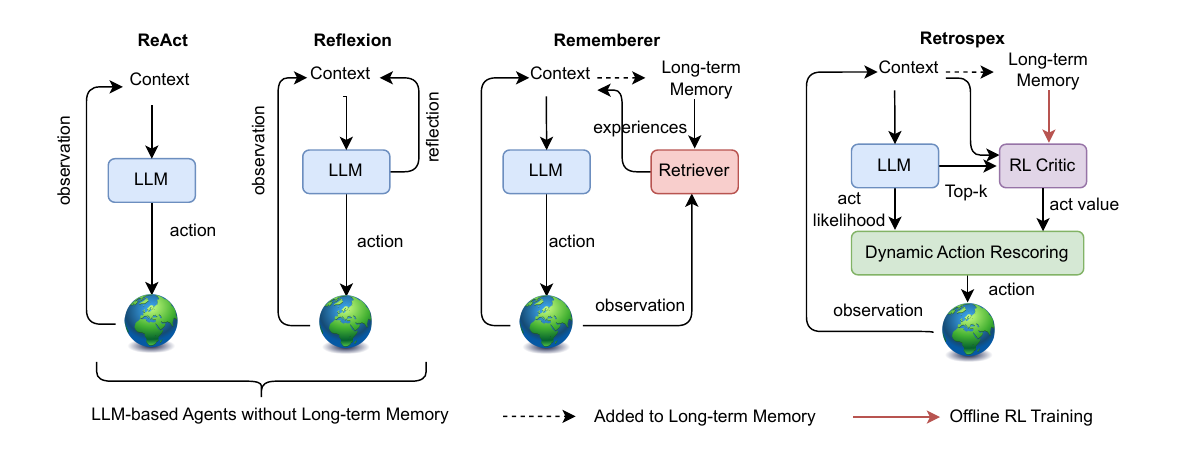}
    \caption{Comparing different architectures for LLM-based Agents}
    \label{LLM-based Agents}
\end{figure*}

In this work, we propose a novel LLM-based agent framework called \modelname{}, which collects cross-task experiences for training a Reinforcement Learning (RL) Critic in a ``retrospection'' stage. The RL Critic is then used to support the LLM in decision making. Unlike previous studies (see Figure \ref{LLM-based Agents}), \modelname{} does not directly integrate experiences into the context. Instead, it exploits an action rescoring strategy to combine the likelihood of the actions provided by the LLM and the action values estimated by the RL Critic. In addition, \modelname{} dynamically increases the weight of action values from the RL Critic for tasks that require more interaction steps with the environment, allowing experiences to gradually play a more important role in difficult tasks.


\modelname{} has several advantages over previous approaches. First, compared to RL-based agents, \modelname{} can leverage the strength of LLMs for more effective decision making. Second, compared to previous LLM-based agents, \modelname{} can better utilize experiences without increasing the context length. Third, \modelname{} is more flexible in controlling how much experience is needed at each step thanks to the dynamic scoring method. Finally, \modelname{} is general and can be adapted to various LLMs or RL methods. In this paper, we implement RL Critic with a lightweight neural network, thus providing little inference overhead compared to using only LLMs for action selection.

We evaluate \modelname{} in three text-based simulation environments: ScienceWorld \cite{wang2022scienceworld}, ALFWorld \cite{shridhar2020alfworld}, and Webshop \cite{yao2022webshop}. The experimental results demonstrate that integrating the RL Critic and dynamic action scoring in \modelname{} enhances the performance of LLM-based agents, leading to success rate improvements of 9\% in ScienceWorld, 3.5\% in ALFWorld, and up to 5\% in Webshop.

Our contributions are summarized as follows:
\begin{itemize}
    \item We propose \modelname{}, a general framework for LLM-based agents that exploits the experience memory for training an RL Critic to support LLMs in decision making.
    
    \item We propose a dynamic action rescoring that combines LLM's likelihood and RL Critic action values. By doing so, we balance the two factors in planning, the current task information and past experiences. 
    
    \item We test our method in three different environments, ScienceWorld, ALFWorld and Webshop. The results are promising and validate our framework's effectiveness. 
\end{itemize}

\section{Related Work}
\subsection{Reasoning and Planning with LLM}
LLMs have been exploited to tackle a wide range of tasks such as reasoning \cite{wei2022chain, kojima2022large, yao2024tree}, self-verification \cite{wang2022self, miao2023selfcheck}, problem decomposition or formalization \cite{madaan2024self, zhou2022least}, and planning \cite{yao2022react, wu2023plan, wang2023describe}.



Most of these aforementioned studies, however, do not utilize the agent's past experiences for performance improvement. To overcome this issue, recent studies leverage relevant experiences to prompt LLM for reasoning, allowing LLM-based agents to learn from previous mistakes. Notable examples include Relexion \cite{shinn2023reflexion}, Rememberer \cite{zhang2024large}, Salam \cite{wang2023learning} and ExpeL \cite{zhao2024expel}. However, this approach is still limited by the context length of LLMs, hindering the ability to fully utilize past experiences. Our proposed approach combines LLM's action likelihood with RL Critic's action values for action reranking. By doing so, there is no need to incorporate experiences into the LLM context, thereby mitigating the problem of long context.

\subsection{LLM combined with RL}

RL has traditionally been used to train agents capable of making sequential decisions. With the advent of large language models (LLMs), many efforts have emerged to integrate LLMs with RL for agent training. These approaches can be broadly categorized into two groups, as outlined below.

The first group uses RL techniques to train LLMs as policy models for acting in new environments. This includes GPT-Critic \cite{jang2022gpt}, LID \cite{li2022pre}, AgentTuning \cite{zeng2023agenttuning}, PAC \cite{springenberg2024offline}, and \att{} \cite{a3t}. LID uses LLMs for policy initialization, while AgentTuning applies imitation learning (IL) to train adaptable agents. GPT-Critic and PAC, on the other hand, train LLMs as both critics and actors using offline RL. Like AgentTuning, \att{} and LID, we use IL to train a base LLM for decision-making. However, unlike these methods, \modelname{} also focuses on enhancing the base LLM's performance during inference without LLM update. This approach avoids the computational cost and potential risk of weakening the LLM's general capabilities that could arise from frequent updates.

The second group uses RL methods to train assistants that support LLMs in decision-making via prompting, including Salam \cite{wang2023learning}, SayCan \cite{brohan2023can}, and Rememberer \cite{zhang2024large}. Salam uses IL to train an assistant who corrects mistakes and provides guidelines. Rememberer uses Q-learning to estimate action values for past experiences stored in memory. During inference, Rememberer retrieves the most relevant experiences (along with their corresponding action values) and incorporates them into the LLM's context. Both Salam and Rememberer extend the LLM's context with additional information. SayCan, on the other hand, estimates an affordance function to help ground the LLM's actions. The final action probability is calculated by combining the LLM's likelihood with the affordance value. 

Our approach is most closely related to SayCan but differs in two key aspects: 1)  SayCan's affordance function is used for action grounding, whereas RL Critic in \modelname{} is for action re-evaluation. SayCan allows for combining any LLM with any affordance function, even independently trained ones. In contrast, we train an RL Critic on the action distribution supported by the LLM, enabling better value estimates for LLM's actions; 2) \modelname{} exploits dynamic scoring, whereas SayCan employs a static score combination.

\section{Methodology}
\begin{figure*}[]
    \centering
    \includegraphics[width=0.95\textwidth]{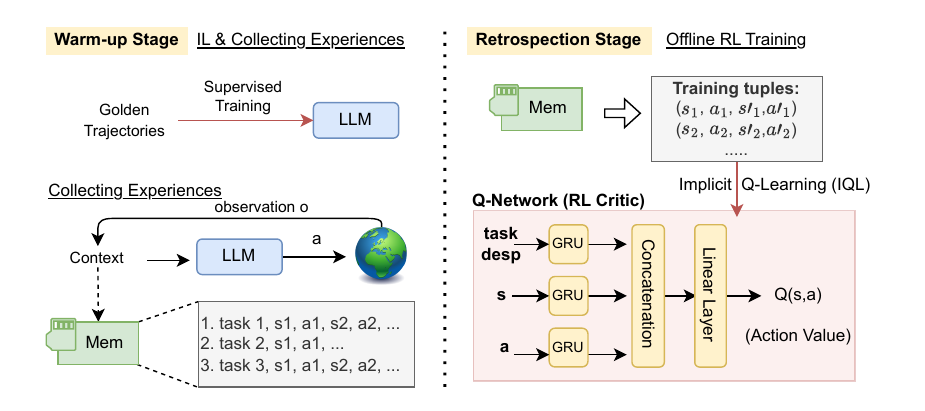}
    \caption{The training process of \modelname{}  includes two stages: 1) In the \textbf{Warm-up stage}, an imitation learning (IL) base agent is trained and used for experience collection; 2) In the \textbf{Retrospection stage}: Offline RL is used to train RL Critic from offline experiences. Here, $s*$ and $a*$ denote states and actions, respectively. In the retrospection stage, $s'$ and $a'$ indicate the following state and action.}
    \label{Retrospex-training}
\end{figure*}

Figure \ref{Retrospex-training} demonstrates the training process of \modelname{} which involves a warm-up phase and a retrospection phase. During the warm-up phase, we fine-tune the LLM based on expert (e.g. human) demonstrations and collect the working experiences of the LLM agent. In the retrospection stage, we train an RL Critic from the LLM agent's experiences using Implicit Q-learning (IQL), an offline RL method. By doing so, the RL Critic is expected to learn from the LLM agent's mistakes and assist in making better decisions in the future.


\subsection{Warm-up Stage}

\paragraph{Imitation Learning} 
Inspired by \cite{lin2024swiftsage, zeng2023agenttuning}, we cast the action prediction task as a text generation task. We fine-tune the LLM with expert demonstrations (i.e., golden trajectories). The objective is to equip the LLM with fundamental knowledge about the agent's environment. This process, known as Imitation Learning (IL), is essential for LLMs of moderate size but can be skipped for powerful LLMs such as GPT-4. 

Formally, we train the LLM policy $y=\pi(x)$ so that the generated action $y$ is the most likely action $\pi^*(x)$ taken by a human expert. Here, $x$ is the given context that contains \textit{a task description}, and a sequence of \textit{states}, and \textit{actions}. A \textit{state} encapsulates the environment information at a specific time. For simplicity, we assume that \textit{states} can be inferred from the initial agent state and subsequent observations from the environment. A \textit{golden trajectory} $\xi=\{task1, s_1, a_1, s_2, a_2, s_3, a_3\}$  can be decomposed into multiple (training) instances ($x_1=\{task1, s_1\}$, $y_1=a_1$), ($x_2=\{task1, s_1, a_1, s_2\}$,\\ $y_2=a_2$), and ($x_3=\{task1, s_1, a_1, s_2, a_2, s_3\},\\ y_3=a_3$). The training objective then involves solving the following optimization problem:
\begin{equation}
\hat{\pi}^{*}_{LLM}=\underset{\pi}{\arg \min } \sum_{\xi \in \mathcal{T}} \sum_{x \in \xi} L_{NLL}\left(\pi(x), \pi^{*}(x)\right) \nonumber
\end{equation}

\noindent where $\mathcal{T}$ denotes the set of golden trajectories and $\xi$ is one particular trajectory. $L_{NLL}$ represents the negative log-likelihood loss, $\pi^{*}(x)$ denotes the expert's action for state $x$, and $\hat{\pi}^{*}_{LLM}$ is the estimated policy model (the fine-tuned LLM).


\paragraph{Collecting Experiences} Due to the complexity of the environment and the limited size of demonstration data, IL is often insufficient for obtaining an optimal policy. 
As such, we collect the experiences of the trained LLM interacting with the environment. Here, the format of each experience trajectory is similar to that of a golden trajectory, but an experience may contain suboptimal actions and/or be a failed attempt to finish users' tasks.

\subsection{Retrospection Stage} 
The task of sequential decision can be formalized as a Markov Decision Process (MDP), which is denoted as $\left(\mathcal{S}, \mathcal{A}, p_{0}(s), p\left(s^{\prime} \mid s, a\right), r(s, a), \gamma\right)$. Here, $\mathcal{S}$, $\mathcal{A}$ are the state and action spaces, $p_{0}$ is the initial state distribution, $p\left(s^{\prime} \mid s, a\right)$ is the environment dynamics, $r(s, a)$ is a reward function, and $\gamma$ is a discount factor. The objective is to find a policy $\pi(a|s)$ that maximizes the cumulative discounted return as follows:
\begin{equation}
\begin{split}
    \pi^{*} &=\underset{\pi}{\arg \max } \mathbb{E}_{\pi}[\sum_{t=0}^{\infty} \gamma^{t} r\left(s_{t}, a_{t}\right) \\
    & \mid s_{0} \sim p_{0}, a_{t} \sim \pi\left(\cdot \mid s_{t}\right), s_{t+1} \sim p\left(\cdot \mid s_{t}, a_{t}\right) ]\nonumber
\end{split}  
\end{equation}

\begin{figure*}[]
    \centering
    \includegraphics[width=0.95\textwidth]{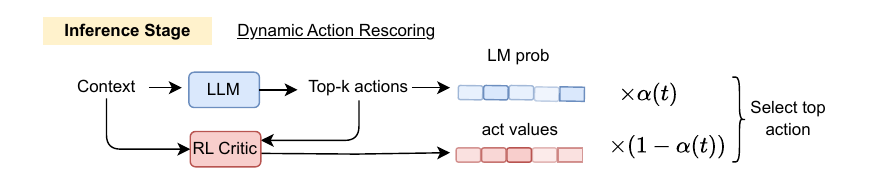}
    \caption{Dynamic Action Rescoring in \modelname{}, where $t$ indicates the interaction step in the current trajectory.}
    \label{Retrospex-testing}
\end{figure*} 

RL can be used to solve the MDP problem and find $\pi$ using interaction data. In general, RL can be conducted online, where we update the LLM-based agent whenever we have a new experience. However, doing so can be expensive and unstable \cite{nottingham2023selective}. As a result, we follow the offline RL approach, where we collect experiences to memory, and update the LLM-based agent once we have enough experience. 

Offline RL uses a fixed experience memory to train the action-value function $Q(s, a)$. Here, the $Q$ value corresponds to the expected cumulative reward (return) obtained by starting from the state $s$, performing action $a$, and then following the policy $\pi$. This work exploits \textit{Implicit Q-Learning} (IQL) \cite{kostrikov2021offline}, which aims to handle the \textit{issue of overestimating Q-function due to unseen actions in offline RL}. 
IQL builds on approximate dynamic programming methods that minimize temporal difference errors as follows:
\begin{equation}
\begin{split}
    L_{T D}(\theta) &= E_{(s, a, s^{\prime}) \sim \mathcal{D}}[(r(s, a)+\\
    & \gamma \max _{a^{\prime}} Q_{\hat{\theta}}(s^{\prime}, a^{\prime})-Q_{\theta}(s, a))^{2}]
\end{split} \nonumber
\end{equation}

Here, $D$ is the experience memory, $s, s'$ are the current and next states respectively. A target Q-network $Q_{\hat{\theta}}$ is used for action selection, and an online Q-network $Q_{\theta}$ is used for value estimation update at each training step. After each training batch, the target network is updated based on the online network.
To prevent the target network from selecting actions that are not supported in the experience memory (due to $\max _{a^{\prime}} Q_{\hat{\theta}}$) , IQL applies a separate state value function $V(s)$ to estimate:
\begin{equation}
    L_{V}(\phi) = E_{(s,a)\sim D}[L_{2}^{\tau}(Q_{\hat{\theta}}(s,a)-V_{\phi}(s))] \nonumber
\end{equation}

Let $u=Q_{\hat{\theta}}(s,a)-V_{\phi}(s)$, $L_{2}^{\tau}(u)=|\tau-\mathbbm{1}(u<0)| u^{2}$ is the upper expectile function. It has been proven that, by optimizing the above objective, we \textit{fit $V_{\phi}(s)$ to approximate the maximum of $Q_{\hat{\theta}}$ over actions supported by the data distribution when $\tau\to 1$} (Theorem 3 by \citet{kostrikov2021offline}). After this estimation, IQL can apply $V(s)$ to update the $Q(s,a)$ with simple MSE loss.
\begin{equation}
    L_{Q}(\theta) = E[(r(s,a)+\gamma V_{\phi}(s^{'})-Q_{\theta}(s,a))^{2}] \nonumber
\end{equation}
where the expectation is calculated by sampling $(s,a,s^{'})\sim D$. The value functions (Q-network and V-network) can be realized in many forms, here we use GRU neural networks as shown in Figure \ref{Retrospex-training}. We encode task description, state, and action separately with different GRU blocks, then concatenate the embeddings together and send them to the next linear layers. We use 2 linear layers after the encoding layer to get the final $q$ and $v$ values. The structure of the V-network is similar to Q-network except that we do not have action $a$ as the input and the output is the state value $V(s)$.

\subsection{Inference Stage}
After training, LLM and RL Critic (Q network) are used for inference as follows.
\paragraph{Action Mapping} 
To ensure that the action candidates sampled by the LLM can be executed in the environment, we map the LLM response to the valid action space as follows: 
\begin{enumerate}[label=\alph*)]
    \item \label{1)} Given the current context including task description, past states and actions, the LLM generates $K$ next action candidates using nuclear sampling with a temperature of 1 and top-p of 0.95;
    \item \label{2)}Filter out the candidates that are already valid actions (assuming there are $m$ candidates), and directly retain this part of the candidates;
    \item \label{3)} Obtain embeddings of the $(K - m)$ invalid candidates, and those of all valid actions through the Sentence Transformer \cite{reimers2019sentence}; Find the $(K - m)$ valid actions that have the largest sum of cosine similarities with all the candidates, excluding those that have been included in Step \ref{2)};
    \item \label{4)} The final candidate of all actions is the union of actions in Step \ref{2)} and Step \ref{3)}.
\end{enumerate}

\paragraph{Dynamic Action Rescoring}
The LLM first generates several responses by nuclear sampling. After mapping the responses into action space as aforementioned, LLM provides the probabilities $p$ of these action candidates. We then normalize these values to obtain LLM scores. 
\begin{equation}
    p = Norm(\hat{\pi}^{*}_{LLM}(a|s)) \nonumber
\end{equation}

\noindent The top-K actions are fed into the RL Critic. The value function will give the action value $q$ for each action, which is then normalized as follows. 
\begin{equation}
    q = Norm(Q_{IQL}(s, a)) \nonumber
\end{equation}
To decide which candidate will be chosen, we combine the 2 scores together as the final score $S$ and select one with the highest score as follows:
\begin{equation}
    \alpha(t) = max(b, d^{t}) \nonumber
\end{equation}
\begin{equation}
    S(a) = \alpha(t)p + (1-\alpha(t))q \nonumber
\end{equation}
where $b,d\in[0,1]$ and $\alpha(t)$ is the dynamic combination weight between $p$ and $q$. When $t=0$ (corresponding to $\alpha(t)$ of $1$), we have few observations and rely more on the commonsense of the LLM to guide action selection. As $t$ increases, $\alpha(t)$ decreases with a discount factor $d$, allowing the RL Critic to have a greater influence on decision-making. However, we set a lower bound of $b$ for $\alpha(t)$ to ensure the LLM’s role is not reduced too drastically for long trajectories.  

\begin{table*}[]
\centering
\begin{tabular*}{0.95\linewidth}{c|cc|ccc}
\toprule
  \multirow{2}{*}{Env}         & \multicolumn{2}{c|}{\textbf{Warm-up Stage}}                                                     & \multicolumn{3}{c}{\textbf{Retrospection Stage}} \\ \cmidrule{2-6}
  \rowcolor{my_gray}            & LLM & \# Training data & IQL   & \# Training Trajs   & SR  \\ \midrule
SciWorld & Flan-T5-large (770M)                      & 2157   golden                                       & GRU (2.7M)  & 2566             & 36  \\
 ALFWorld & \multirow{2}{*}{LLaMA3-8B-Instruct} & \multirow{2}{*}{AgentInstruct+} & GRU (2.7M)  &   2000              & 67  \\
Webshop &                                     &                                           & GRU (2.7M)  &   1500              & 44  \\ \bottomrule
\end{tabular*}
\caption{Training data used in the warmup and retrospection stages of \modelname{}. Here, AgentInstruct+ is a dataset used by \cite{zeng2023agenttuning} for agent training, which consists of 1866 golden trajectories from a mix of 6 environments including Webshop (351 golden trajectories) and ALFWorld (336 golden trajectories). SR denotes the percentage of the successful trajectories in the memory used for retrospection stage training. We train one LLM for both ALFWorld and Webshop environments.}
\label{tab:all-training}
\end{table*}

\begin{table}[]
\centering
\begin{tabular}{ccc}
\toprule
\textbf{Env} & \# \textbf{Subtasks} & \# \textbf{Test Samples} \\ \midrule
SciWorld&30&270     \\
ALFWorld&6&134 \\
Webshop&-&100 / 200 / 251\\
\bottomrule
\end{tabular}
\caption{The number of subtasks and testing samples in the three tested environments. For Webshop, we conduct evaluation on three different test sets used by \citet{zhang2024large} (100 samples), \citet{zeng2023agenttuning} (200 samples), and \citet{a3t} (251 samples).}
\label{tab:test details}
\end{table}


\section{Experiments}
The experiments are conducted in three environments: ScienceWorld \cite{wang2022scienceworld}, ALFWorld \cite{shridhar2020alfworld} and Webshop \cite{yao2022webshop}. We use Average Score (AS) and Success Rate (SR) to measure the performance\footnote{ALFWorld only supports SR for evaluation}. Both metrics (AS and SR) are scaled to the range of $[1,100]$ in all three environments. The training and testing sets for the three environments are summarized in Table \ref{tab:all-training} and Table \ref{tab:test details}. 

\subsection{ScienceWorld}
\paragraph{Experimental Setup} ScienceWorld is a complex interactive text environment that tests an agent's scientific commonsense. In this environment, agents are required to navigate through 10 interconnected locations (e.g., workshop, kitchen) and utilize the tools to complete tasks such as “determine if the metal fork is electrically conductive”. The environment contains 200 objects, 25 high-level actions, resulting in approximately 200k possible action-object combinations. We collect 2157 golden (i.e. successful) trajectories to train Flan-T5-large\footnote{https://huggingface.co/google/flan-t5-large} in the warm-up stage, and 2566 trajectories, which contain both fail and successful ones with SR of 36\%, for training GRU-based RL Critic in the retrospection stage. In the warm-up stage, we follow the same training strategy for imitation learning (IL) specified in SwiftSage \cite{lin2024swiftsage}. We denote this IL agent as IL-T5, which corresponds to the Swift model in SwiftSage as well as Retrospex (w/o retrospection). For dynamic re-scoring, we choose $d=0.97,b=0.6$ as our hyper-parameters.


\paragraph{Baselines} We compare Retrospex to baselines of different types: (1) LLM-based agents ReAcT \cite{yao2022react} and Reflexion \cite{shinn2023reflexion}; (2) online RL agent DRRN \cite{he2016deep}; and (3) SayCan \cite{brohan2023can} which combines LLM with an affordance function for action grounding. The details for SayCan, ReAct and Reflexion are provided in SwiftSage paper \cite{lin2024swiftsage}. It is noteworthy that we do not compare to SwiftSage here as it exploits two large language models (GPT and IL-T5) for inference, resulting in a somewhat unfair comparison. The results of GPTP3.5-based ReAct and DRRN are produced by ourselves.


\begin{table}[]
\centering
\scalebox{0.95}{
\begin{tabular}{cccc}
\toprule
\textbf{Method} & \textbf{LLM} & \textbf{AS}   & \textbf{SR}  \\ \midrule
DRRN  (our run)               & GRU                    & 14.13        & 0                \\
ReAcT (our run)                  & GPT3.5                   & 14.60        & 0.03               \\
SayCan \dag               & GPT4+SBERT             & 33.82         & -                \\ 
ReAcT \dag                  & GPT4                   & 36.43        & -                \\
Reflexion \dag          & GPT4 & 45.34 &-\\
IL-T5                     & Flan-T5-large          & 48.80        & 27.0            \\
\rowcolor{my_gray}Retrospex    & Flan-T5-large          & \textbf{55.98}         & \textbf{36.0}             \\ \bottomrule
\end{tabular}}
\caption{The AS and SR on ScienceWorld. Here, {\dag} denotes the results from SwiftSage \cite{lin2024swiftsage}.} 
\label{tab:all-sci}
\end{table}

\paragraph{Results} The experimental results on ScienceWorld are shown in Table \ref{tab:all-sci}, where several observations can be drawn. Firstly, it can be seen that the performance of DRRN is significantly worse than other baselines, confirming \textit{the challenge of learning an independent RL agent in an environment with a large action space}. Secondly, GPT4-based ReAct models can achieve relatively good results without any training, suggesting that \textit{powerful LLMs can exploit its commonsense to recognize meaningful action-object combinations} for better results. Thirdly, the result of SayCan is not satisfactory with GPT4+SBERT, suggesting that exploiting \textit{a value function that is trained independently from the LLM-based actor might not be optimal}. Last but not least, it is observable that IL-T5, a relatively small-size LLM-based IL agent, can outperform Reflexion, which is based on the powerful GPT4 model. This shows that the knowledge of small size LLM (T5) might be sufficient for ScienceWorld, and IL is important to ground an agent in the targeted environment. In addition, the fact that \modelname{} significantly outperforms IL-T5 by 7 points in AS and 9 points in SR suggests \textit{the importance of the retrospection stage for agents to explore and learn from mistakes}. A more detailed list of all 30 sub-tasks of ScienceWorld can be seen in Table \ref{tab:sci} in the Appendix.

\subsection{ALFWorld}
\paragraph{Experimental Setup} ALFWorld is also a suite of text-based environments that challenge an agent to solve multi-step tasks based on TextWorld \cite{Ct2018TextWorldAL}. The action and task formats of ALFWorld are similar to ScienceWorld but simpler. 
We exploit the AgentInstruct+ dataset for training a LLaMA3-8B-Instruct-based IL agent in the warm-up stage and collect 2000 trajectories with SR of 67\% for restrospection stage training. Here, AgentTuining+ includes  both AgentInstruct\footnote{https://huggingface.co/datasets/THUDM/AgentInstruct} for agent learning capabilities and ShareGPT\footnote{https://huggingface.co/datasets/anon8231489123/Share\\GPT\_Vicuna\_unfiltered} for general capability. The AgentInstruct+ dataset contains the golden trajectories of both ALFWorld and Webshop environments, thus we train only one IL agent for being used in both environments.  The combination parameters of dynamic action rescoring in \modelname{} for ALFWorld are $d=0.95,b=0.6$.

\begin{table}[]
\centering
\scalebox{0.95}{
\begin{tabular}{ccc}
\toprule
\textbf{Method} & \textbf{LLM}  & \textbf{SR}  \\ \midrule
Reflexion  \dag                 & GPT3.5                 & 40.3 \\
ExpeL  \dag        &GPT3.5 & 59.0 \\
$A^3T$ (round=0)                    & Mistral-7B-Instruct                    & 86.0 \\
$A^3T$ (round=1)                    & Mistral-7B-Instruct                    &\textbf{94.0} \\
IL-LLaMA3               & LLaMA3-8B-Instruct                 &83.5\\ 
\rowcolor{my_gray}{Retrospex}               & {LLaMA3-8B-Instruct}          &\textbf{87.0} \\ \bottomrule
\end{tabular}}
\caption{Overall results on ALFWorld. Here, {\dag} denotes the results of Reflexion \cite{shinn2023reflexion} and ExpeL \cite{zhao2024expel} from \cite{zhao2024expel}. The result of $A^3T$ is from \cite{a3t}.} 
\label{tab:alf}
\end{table}

\paragraph{Baselines} We compare to Reflexion (GPT 3.5) and ExpeL \cite{zhao2024expel} and \att{} \cite{a3t}. Concurrent to our work, \citet{a3t} proposes \att{}, a self-improvement framework to train LLM agents by updating LLM multiple rounds. For the first round (round=0), imitation learning is used to train LLM from golden trajectories, which is similar to our warm-up stage. For other rounds (round > 0), \att{} updates LLM using contrastive learning methods like DPO \cite{rafailov2023direct}. \att{} also proposes an interesting method for generating composed trajectories for exploration. The details of Reflexion, ExpeL, $A^3T$ can be found in \att{} paper \cite{a3t}.

\paragraph{Results}

Table \ref{tab:alf} presents the performance of compared methods on the ALFWorld environment. The results indicate that \modelname{} outperforms the base IL model (\illama{}) and \att{} (round=0) but falls short when compared to \att{} (round=1). Upon examining the training details of \att{}, we observe that \att{} leverages 981 golden trajectories for round=0 and 3431 trajectories (with a 90.2\% success rate) for round=1. In contrast, \modelname{} uses only 351 Webshop golden trajectories during the warm-up phase and 2000 trajectories with a lower success rate of 67\% for the retrospection phase (see Table \ref{tab:all-training}). The discrepancy in data quality and quantity accounts for the underperformance of \illama{} compared to \att{} (round=0) and of \modelname{} compared to \att{} (round=1).

Although \modelname{} does not achieve state-of-the-art (SOTA) performance on ALFWorld, our retrospection strategy remains valuable for three key reasons. First, the retrospection phase significantly improves the base IL agent, as evidenced by the 3.5\% success rate (SR) increase of \modelname{} compared to \illama{}. Second, the retrospection phase is much more cost-effective than a full training round in \att{}. Specifically, \att{} requires direct updates to a large (7B) LLM, while \modelname{} only updates a smaller RL-Critic model—a GRU with 2.7M parameters. Frequently updating LLM is not only computationally expensive but also risks weakening its general capability due to catastrophic forgetting. Finally, the strategies of \modelname{} and \att{} can be combined as a small RL-Critic model can be trained to assist with inference between LLM update cycles in \att{}, offering a more practical approach.


\subsection{Webshop}

\begin{table}[]
\centering
\scalebox{0.95}{
\begin{tabular}{cccc}
\toprule
\textbf{Method} &  \textbf{LLM} & \textbf{AS}   & \textbf{SR}  \\ \midrule
\multicolumn{4}{c}{Rememberer \cite{zhang2024large} test set } \\ \midrule
ReAcT \dag   & GPT3.5   & 66.0   & 36.0\\
Rememberer \dag & GPT3.5   & 68.0   & 38.0   \\
IL-LLaMA3  & LLaMA3 8B & 76.2  & 42.4  \\
\rowcolor{my_gray} Retrospex & LLaMA3 8B & 74.6  & \textbf{46.0} \\
\midrule
\multicolumn{4}{c}{AgentLM \cite{zeng2023agenttuning} test set} \\ \midrule
ReAcT \ddag    & GPT4      & 58.6    & -   \\
AgentLM \ddag  & LLaMA2 7B  & 63.6 & -      \\
AgentLM \ddag  & LLaMA2 13B  & 70.8 & -      \\
IL-LLaMA3   & LLaMA3 8B  & 77.1   & 45.5   \\
\rowcolor{my_gray} Retrospex  & LLaMA3 8B & \textbf{77.7}  & \textbf{50.5} \\
\midrule
\multicolumn{4}{c}{AgentBoard \cite{ma2024agentboard} test set } \\ \midrule
$A^3T$ (round=0)   & Mistral 7B   & 72.0   & - \\
$A^3T$ (round=1)   & Mistral 7B   & 73.5   & - \\
IL-LLaMA3    & LLaMA3 8B  & 76.5   & 44.2   \\
\rowcolor{my_gray} Retrospex  & LLaMA3 8B & \textbf{77.2}  & \textbf{49.0} \\
\bottomrule
\end{tabular}}
\caption{Overall results on Webshop, where {\dag} and {\ddag} results are from \cite{zhang2024large} and \cite{zeng2023agenttuning}.  The result of $A^3T$ is from \cite{a3t}.}
\label{tab:all-web}
\end{table}

\paragraph{Experimental Setup}  Webshop \cite{yao2022webshop} is an online shopping website environment with 1.18M products and 12k human instructions. An agent is required to purchase a product based on a user instruction such as ``I am looking for a nightstand with drawers.'' To complete the task, the agent needs to perform actions such as searching ``nightstand drawers,'' or choosing clickable buttons. We use the same  \illama{} as the IL base agent as described in the previous section. The RL-Critic, however, is trained specifically for Webshop. The parameters of dynamic action rescoring for Webshop are $d=0.9,b=0.5$.

\paragraph{Baselines} We compare with three main baselines Rememberer \cite{zhang2024large}, AgentLM \cite{zeng2023agenttuning}, and \att{}. As different studies conduct evaluations on different subsets of the original Webshop test tasks, we conduct multiple tests for fair and comprehensive comparisons. Specifically, Rememberer uses the first 100 samples for testing whereas AgentLM uses 200 samples. \att{} both compares with AgentLM \footnote{\att{} doesn't explicitly state if AgentLM test set is used} and reports the result on AgentBoard test set. Here, we compare to \att{} results on the AgentBoard test set. 

\paragraph{Results} Table \ref{tab:all-web} shows the performance of \modelname{} and other baselines on Webshop environment. The experiment verifies the effectiveness of \modelname{} over Rememberer, AgentLM, and \att{} (round=0), (round=1) on their respective reported test sets. It is noteworthy that \modelname{} performs well compared to \att{} even though we use one base LLM for ALFWorld and Webshop, whereas \att{} finetunes another LLM specifically for Webshop. The retrospection stage in \modelname{} helps improve the success rate (SR) significantly over three test sets and improves AS in two over three test sets. One explanation for why \modelname{} obtains higher SR yet lower AS in the Rememberer test set is that \illama{} may not return correct products in many cases (fail cases) yet it returns close enough products (high scores). We leave further investigation to future work.

\section{Further Analysis}
\subsection{Analysis on Task Complexities}

To evaluate the performance of \modelname{} across varying task complexities, we categorize the tasks in ScienceWorld into three levels: \textit{short} (fewer than 20 steps), \textit{medium} (20 to 50 steps), and \textit{long} (more than 50 steps). We then calculate the average score (AS) for each complexity level, with the results presented in Table \ref{tab:complexity}. Across all three levels, \modelname{} demonstrates significant improvements. Notably, tasks with medium-length trajectories show an average score increase of more than 10 points, while tasks with short and long trajectories see improvements of over 5 points.

\begin{table}
\centering
\begin{tabular}{cccc}
\toprule
          & Short           & Medium          & Long            \\ \midrule
SayCan    & 43.83          & 36.55          & 23.65          \\
IL-T5 & 90.87           & 42.71         & 25.56          \\
T5-then-IQL    & 68.17          & 31.18          & 19.93          \\  
\rowcolor{my_gray} Retrospex & \textbf{95.49} & \textbf{55.18}  & \textbf{31.93} \\ \bottomrule
\end{tabular}
\caption{Average reward scores on different task complexity on ScienceWorld}
\label{tab:complexity}
\end{table}

We also report the results for T5-then-IQL, where the top actions are reranked based solely on RL-Critic scores. The inferior performance of T5-then-IQL compared to IL-T5 suggests that the LLM's likelihood should not be disregarded when selecting actions. This performance drop is especially pronounced in tasks with short trajectories, highlighting the importance of LLM when we have fewer observations. This supports our intuition behind dynamic scoring, where we place greater trust in the LLM when the step count $t$ is small.

\begin{table*}[]
\centering
\begin{tabular*}{0.88\linewidth}{cccccccc}
\toprule
 & IL-T5 & \begin{tabular}[c]{@{}c@{}}$d$=0\\ $b$=0\end{tabular} & \begin{tabular}[c]{@{}c@{}}$d$=0.95\\ $b$=0.25\end{tabular} & \begin{tabular}[c]{@{}c@{}}$d$=0.97\\ $b$=0.5\end{tabular} & \begin{tabular}[c]{@{}c@{}}$d$=0.97\\ $b$=0.6\end{tabular} & \begin{tabular}[c]{@{}c@{}}$d$=0.99\\ $b$=0.6\end{tabular} & \begin{tabular}[c]{@{}c@{}}Static combination\\0.6$p$+0.4$q$\end{tabular} \\ \midrule
Short      & 90.87  & 68.17                                            & 94.63                                                  & \textbf{96.43}                                        & 95.49                                                 & 93.11                                                               & 94.35                                                   \\
Medium     & 42.71   & 31.18                                            & 52.40                                                  & 54.68                                                  & \textbf{55.18}                                         & 51.63                                             & 54.63                                                  \\
Long       & 25.56   & 19.93                                            & 25.58                                                  & 29.33                                                & 31.93                                               & \textbf{35.39}                                        & 29.21                                                  \\ \midrule
AS & 48.80  & 36.7                                              & 52.51                                                  & 55.13                                                 & \textbf{55.98}                                         & 55.63                               & 54.37                                                  \\ \bottomrule
\end{tabular*}
\caption{Results on ScienceWorld with different dynamic scoring parameters}
\label{tab:diff-parameter}
\end{table*}



\subsection{Analysis on Combination Parameters}

In order to investigate the effect of differences in the parameters used for the combination of dynamic action scoring, we test different $d$ and $b$ on ScienceWorld and the results are shown in Table \ref{tab:diff-parameter}.
\paragraph{Necessity of Two Scores}
When using only the IQL score ($d=0,b=0$), the average scores drop significantly for three different complexity settings. When we use LLM only (first column), the performance is poor compared to the results of the score combination. Both scores, therefore, are essential for performance improvement.
\paragraph{Necessity of Dynamic Combination}
We study if a dynamic combination is necessary. For that, we select and fix a combination of the action likelihood and value based on the study of different $b$ values. The last column of Table \ref{tab:diff-parameter} shows the score with a static combination. It is observable that static combination underperforms dynamic combination, verifying the role of the discount factor $d$ in incorporating more experiences for long-horizon tasks.
\paragraph{Parameter Choice}
We analyze the two parameters $d$ and $b$ in dynamic action rescoring. First, when $d$ is small, the weight of the LLM score decreases rapidly within a smaller number of steps. Consequently, the agent quickly shifts the focus to action values from IQL, leading to a drop in performance. Secondly, for our method to be sufficiently effective, the LLM still needs a relatively large weight even at the end of a long trajectory. As such, we need to keep the value of $b$ high to ensure that the validity of the LLM scores is maintained. 


\subsection{Analysis on Inference Time Cost}
Given the context of length $N$ containing past interactions and thoughts, LLM-based agents need to generate the next action. As the action length is often much shorter compared to the context length, we simplify the analysis by estimating the inference time for ReAct, Reflexion, Rememberer and \modelname{} to generate one-token action with Transformer-based LLM.

\paragraph{ReAct} The time for ReAct to generate one-token action is dominated by attention operations in LLM, which is $N^2\times T_1$ where $T_1$ is the computational time depending on the LLM model and the hardware infrastructure. 

\paragraph{Reflexion} The time for Reflexion to generate one-token action is dominated by $(N+R)^2\times T_1$, where $R$ contains the historical trials and errors.

\paragraph{Rememberer} The time complexity is $(N+Kl)^2T_1+ (M T_2+ M\log{K})$. Here, $K$ indicates the number of experiences incorporated into the LLM context, and $l$ is the average experience length. Assuming bruteforce search with the support of a max-heap, $M T_2+M\log{K}$ is the time for retrieving $K$ relevant experiences in the memory of size $M$, and $T_2$ is the time for calculating the similarity between the current trajectory and each trajectory in the memory. The size of memory $M$ will be accumulated, leading to longer inference over time. 

\paragraph{\modelname{}} The time for one-token action generation is $N^2\times T_1+KT_3$ where $T_3$ is the time for calculating the Q-value with GRU. As $T_1\gg T_3$, Retrospex adds little inference overhead compared to ReAct. However, as we can obtain better performance with \modelname{} with smaller LLM (smaller $T_1$), \modelname{} can still win in inference time compared to ReAct based on GPT4. Compared to Reflexion and Rememberer, due to shorter context, Retrospex is more efficient.
In \modelname{}, we still need to sample top-K actions, however, this is done only on the last layer, which is much less demanding compared to $N^2\times T_1$ for LLM.

\section{Conclusion}
This work introduces a novel LLM-based agent framework, named \modelname{}, that addresses the limitations of prior approaches in leveraging experiences for decision-making. \modelname{} overcomes the context length restriction by separating experience analysis (through a Reinforcement Learning Critic) from LLM-based action selection. This enables the agent to effectively utilize past experiences while retaining the strengths of LLMs. Additionally, the dynamic action rescoring method allows for flexible control over the influence of experiences based on task complexity. Evaluations demonstrate that \modelname{} achieves significant performance improvements compared to strong baselines, highlighting its potential for real-world applications of LLM agents.



\section*{Limitations}
There are several limitations to our work. First, LLM-dependent action sampling and short-term evaluation can include limitations and biases from the LLM itself. In addition, the closely related memory can also suffer from distributional biases in trajectories and lack of exploration of the action space. Future work can be investigated to expand the action exploration in the collection stage, thus gathering trajectories with more diversity. Second, current work does not explore the potential of incorporating verbal feedback such as those from Reflexion for better retrospection of past experiences. Third, it will be interesting to improve the dynamic action scoring with an automatic module that decides the weights of experiences instead of relying on predefined hyperparameters.

\section*{Ethic Statement}

In our study, we utilize open simulation environments, ensuring that there is no direct interaction with humans that could potentially cause harm. The training data used in our experiments is sourced from publicly available datasets, all of which are thoroughly cited and referenced in the main text to maintain transparency. By limiting our work to simulated settings and publicly accessible data, we minimize ethical concerns related to privacy, consent, and safety. However, it is important to emphasize that while the current study avoids real-world risks, the broader application of LLM-based agents requires careful consideration. As these technologies are increasingly deployed in real-world settings, it is essential to ensure that they are aligned with human values, respect ethical guidelines, and mitigate potential biases. 

\section*{Acknowledgement}
We thank the anonymous reviewers for their constructive feedback that helps improve this work significantly. Our study was partially supported by computing resources from NJU, the State Key Laboratory for Novel Software Technology, and Intelligent Integration Co. LTD (INT2), Vietnam.

\bibliography{acl_latex}

\appendix

\section{Supplementary Details for Dynamic Rescoring Method}
Our method merges the probability of LLM and the Q value from the IQL together and selects the final action. For detail, the LLM first generates several responses by nuclear sampling. After mapping the responses into action space as aforementioned, LLM provides the probabilities $p$ of these action candidates. We then normalize these values to obtain LLM scores. Here $p$ means the probabilities of all actions given the current state. 
\begin{equation}
\begin{split}
    & p = \{\hat{\pi}^{*}_{LLM}(a_i|s)|i=1,2,...,K\}\\
    & Norm(p_i) = \frac{p_i-min(p)}{max(p)-min(p)}
\end{split}
\end{equation}



\noindent The top-k actions are fed into the RL Critic. The value function will give the action value $q$ for each action, which is then normalized as follows. Here $q$ means the q values of all actions at the current state. When the q values are equal to each other, we give a score of 0.5 to all actions.
\begin{equation}
\begin{split}
    & q = \{Q_{IQL}(s, a_i))|i=1,2,...,K\}\\
    & Norm(q_i) = \frac{q_i-min(q)}{max(q)-min(q)}
\end{split}
\end{equation}

\begin{figure}
    \centering
    \includegraphics[width=0.95\linewidth]{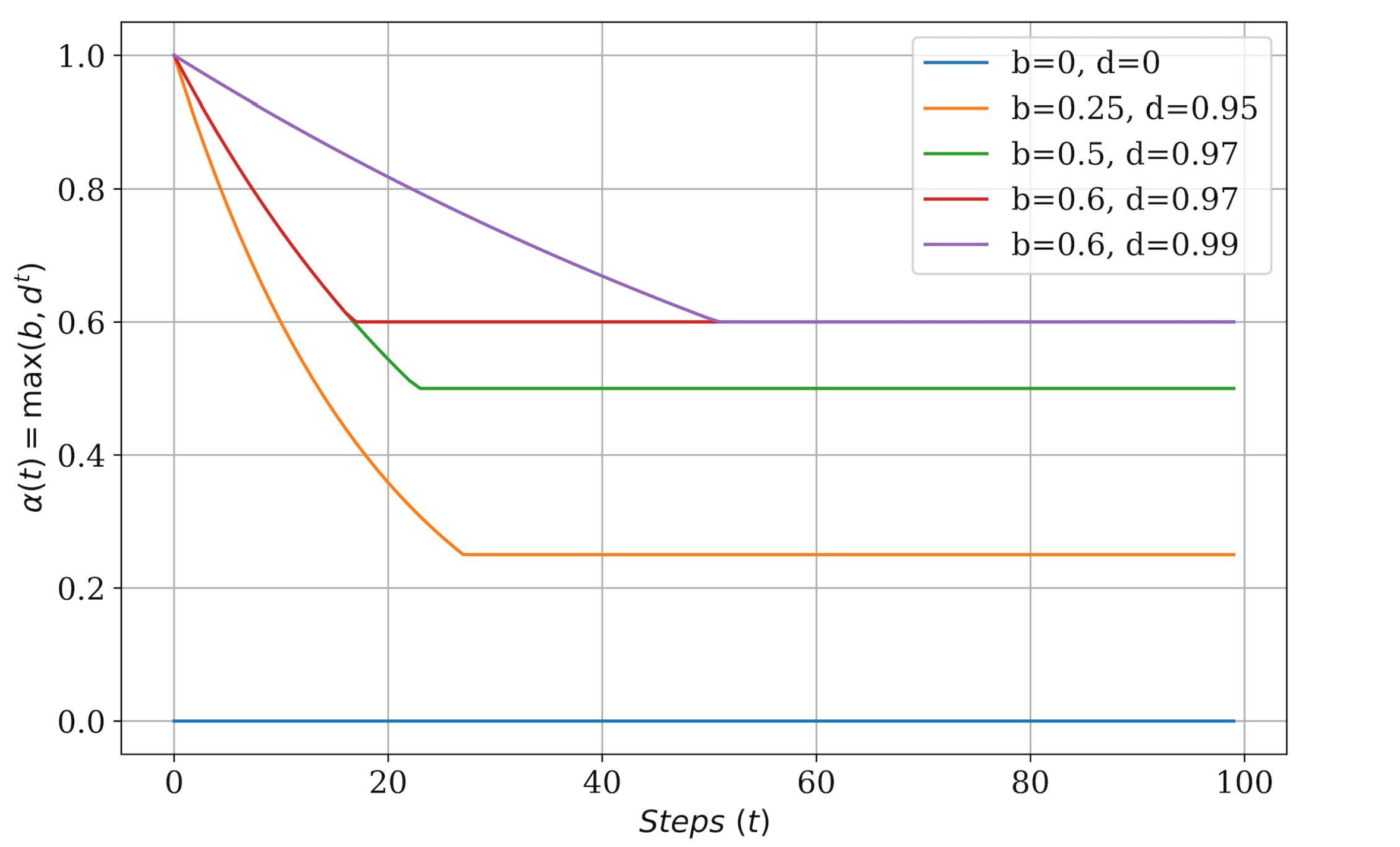}
    \caption{$\alpha(t)$ with different values of steps $t$}
    \label{bd}
\end{figure}

\noindent To decide which candidate will be chosen finally, we combine the 2 scores together as the final score $S$ and select one with the highest score as follows:

\begin{equation}
    \alpha(t) = max(b, d^{t})
\end{equation}
\begin{equation}
    S(a) = \alpha(t)p + (1-\alpha(t))q
\end{equation}
where $\alpha(t)$ is the dynamic combination weight between $p$ and $q$ which changes with different values of the step $t$. In the first step $t=0$, $\alpha(t)$ is $1$ and the agent trusts the trained LLM completely. This is because when the trajectory is short with few observations from the environment, the LLM agent requires fewer experiences for decision. As the step $t$ increases, $\alpha(t)$ will decline with a discount factor $d$, giving more chance for RL Critic to influence the decision making. However, we set the lower bound limit for $\alpha(t)$ to be $b$ so the weight of $p$ will not be too low. The effects of different settings for $b$ and $d$ are shown in Figure \ref{bd}.

\section{Experimental Details}
\label{sec:appendix}

\subsection{Environment Details}
\label{prompts}
Samples of the initial prompts for three environments are listed in the following. In general, each prompt contains an environment description and a task description. The prompt formats of ALFWorld and Webshop are derived from the AgentInstruct dataset \cite{zeng2023agenttuning}. 

\paragraph{\torevise{ScienceWorld Sample Task}}

\paragraph{Task Description} Your task is to find the animal with the longest life span. The animals are in the 'outside' location. Focus on the animal with the longest life span.

\paragraph{Status} Time: 1; Score: 0; 

\paragraph{Action history:} |look around (+0) --> N/A |

\paragraph{Current environment:} This room is called the workshop. In it, you see: | the agent | a substance called air | a table. On the table are: a battery, a black wire, a orange light bulb, which is off, a orange wire, a red wire, a switch, which is off, a violet light bulb, which is off, a yellow light bulb, which is off. | a ultra low temperature freezer. The ultra low temperature freezer door is closed. | You also see: | A door to the hallway | 

\paragraph{Current inventory:} In your inventory, you see: | an orange | 

\paragraph{Visited rooms:} workshop

\paragraph{Question:} What action should you do next?

\paragraph{\torevise{Webshop Sample Task}}

\paragraph{Environment Description} You are web shopping. I will give you instructions about what to do. You have to follow the instructions. Every round I will give you an observation and a list of available actions, you have to respond an action based on the state and instruction. You can use search action if search is available. You can click one of the buttons in [clickables]. An action should be of the following structure: 
\begin{itemize}
    \item search[keywords]
    \item click[value]
\end{itemize}
If the action is not valid, perform nothing. Keywords in search are up to you, but the value in click must be a value in the list of available actions. Remember that your keywords in search should be carefully designed. Your response should use the following format:
\text{Thought: I think ... Action:search/click[...]}
\paragraph{Task Description} Find me machine wash men's dress shirts with cotton spandex, classic fit, short sleeve with color: black, and size: 5x-large tall, and price lower than 60.00 dollars \text{ [Search]}

\paragraph{\torevise{ALFWorld Sample Task}}

\paragraph{Environment Description} Interact with a household to solve a task. Imagine you are an intelligent agent in a household environment and your target is to perform actions to complete the task goal. At the beginning of your interactions, you will be given the detailed description of the current environment and your goal to accomplish. For each of your turn, you will be given a list of actions which you can choose one to perform in this turn. 

\paragraph{Actions} You should choose from two actions: ``THOUGHT'' or ``ACTION''. If you choose ``THOUGHT,'' you should first think about the current condition and plan for your future actions, and then output your action in this turn. Your output must strictly follow this format:
\begin{itemize}
    \item \text{THOUGHT: your thoughts.}
    \item \text{ACTION: your next action}
\end{itemize}

\noindent If you choose ``ACTION'', you should directly output the action in this turn. Your output must strictly follow this format:
\begin{itemize}
    \item \text{ACTION: your next action}. 
\end{itemize}

After your each turn, the environment will give you immediate feedback based on which you plan your next few steps. if the environment output "Nothing happened", that means the previous action is invalid and you should try more options.

\paragraph{Reminder:} 
\begin{enumerate}
    \item The action must be chosen from the given available actions. Any actions except provided available actions will be regarded as illegal.
    \item Think when necessary, try to act directly more in the process. 
\end{enumerate}
    
\paragraph{Initial Observation} You are in the middle of a room. Looking quickly around you, you see a cabinet 16, a cabinet 15, a cabinet 14, a cabinet 13, a cabinet 12, a cabinet 11, a cabinet 10, a cabinet 9, a cabinet 8, a cabinet 7, a cabinet 6, a cabinet 5, a cabinet 4, a cabinet 3, a cabinet 2, a cabinet 1, a coffeemachine 1, a countertop 2, a countertop 1, a diningtable 1, a drawer 5, a drawer 4, a drawer 3, a drawer 2, a drawer 1, a fridge 1, a garbagecan 1, a microwave 1, a safe 1, a sinkbasin 1, a stoveburner 4, a stoveburner 3, a stoveburner 2, a stoveburner 1, and a toaster 1.

\paragraph{Task Description} Your task is to: put a clean spoon in diningtable.

\subsection{ScienceWorld Experimental Details}

\label{impletation details}
\paragraph{Warm-up Stage}

Following the training approach outlined in \cite{lin2024swiftsage}, we enhance the traditional one-hop imitation learning data to multi-hop data by incorporating a sliding window that captures states and rewards from the previous 10 actions (K = 10). Additionally, we introduce a dedicated field to track visited rooms, ensuring no duplication occurs. This approach provides agents with an extended context, thereby preventing redundant room navigation. The main idea is to exploit negative log-likelihood (NLL) loss to train the model to imitate the golden action.

Our backbone model is Flan-T5-large, which is trained with a learning rate of 1e-4 and a batch size of 8. We terminate our model at step 8000. For efficient training, we also employ DeepSpeed Zero-3 \cite{rajbhandari2021zeroinfinity} for parallel training across four V100 GPUs. 

\begin{table*}
\centering
\scalebox{0.90}{
\begin{tabular*}{0.83\linewidth}{cc|ccccc}
\toprule
task    & average \#steps & DRRN   & ReAcT  &SayCan & IL-T5 & Retrospex \\ \midrule
0       & 107.7(L)           & 0      & 0    & \textbf{33.06}      & 29.89    & 14.33      \\
1       & 75.2(L)            & 2      & 0    & 0.37     & 0     & 0      \\
2       & 33.6(M)            & 0      & 0    & \textbf{47.81}      & 29      & 25    \\
3       & 15.1(S)            & 0      & 0     & 39.26      & 28.89    & \textbf{51.11}     \\
4       & 23(M)              & 6.5    & 38.8   & 19.72     & \textbf{44.78}     & 40.44     \\
5       & 14.6(S)            & 4.8    & 18     & 22.87    & 93.2     & \textbf{100}      \\
6       & 14.6(S)            & 5.7    & 17.6  & 31.43  & 93.2      & \textbf{100}     \\
7       & 8.8(S)             & 13     & 0     & 58.18 & \textbf{100}    & \textbf{100}  \\
8       & 12.6(S)            & 10     & 8.6   & 20.87 & 96.6     & \textbf{98.3}      \\
9       & 88.9(L)            & 6      & 19.4  & 3.88   & \textbf{29.78}       & 9.56         \\
10      & 79.6(L)            & 10     & 9     & 13.93 & \textbf{33.8}       & 28.7      \\
11      & 69.5(L)            & 22.6   & 14.9  & 9.92 & 11.2        & \textbf{26.5}     \\
12      & 40(M)              & 17.8   & 10.1  & 20.91 & 24.8      & \textbf{26.6}      \\
13      & 16.3(S)            & 33.6   & \textbf{68.3}  & 16 & 15     & 20.25   \\
14      & 97(L)              & 18.5   & 11.6  & 21.94 & 34   & \textbf{39}      \\
15      & 84.9(L)            & 12.44  & 7.2   & 32.26 & 49.5    & \textbf{51.5}  \\
16      & 123.1(L)           & 7.3    & 5     & 13.67 & 28    & \textbf{54}     \\
17      & 7(S)               & 15.75  & 23    & 80 & \textbf{100}   & \textbf{100}       \\
18      & 8(S)               & 26.67  & 16.67 & 50 & \textbf{100}   & \textbf{100}        \\
19      & 7(S)               & 10.33  & 4.1   & 67.5 & \textbf{100}    & \textbf{100}       \\
20      & 35.2(M)            & 18.17  & 50    & 8.03 & 11.1     & \textbf{65}       \\
21      & 65(L)              & 33     & 30    & 17.41 & 6.8    & \textbf{87.6}       \\
22      & 78.6(L)            & \textbf{50}     & 42.5  & 10.39 & 40     & 28.22      \\
23      & 130.1(L)           & 21     & 0.8   & \textbf{67.53} & 26.5     & 17.8      \\
24      & 132.1(L)           & 20     & 8     & \textbf{59.45} & 17.2    & 25.9      \\
25      & 13.6(S)            & 10     & 4     & 52.14 & \textbf{88.6}  &85.2      \\
26      & 20.8(M)            & 10     & 13.5  & 22.5 & \textbf{62.4}   & 58    \\
27      & 25.6(M)            & 10     & 14.5  & \textbf{99.56} & 60.2    & 69.2     \\
28      & 29(M)              & 16.9   & 1.9   & 47.76 & \textbf{77.8}    & 76.1     \\
29      & 21.4(M)            & 11.9   & 0.7   & 26.37 & 31.6   & \textbf{81.1}      \\ \midrule
Average & -               & 14.13 & 14.6  & 33.82 & 48.80 & \textbf{55.98}    \\ \bottomrule
\end{tabular*}}
\caption{Overall experiment results on ScienceWorld. The result is the average of final scores in 100 steps per trajectory. The result of SayCan comes from \cite{lin2024swiftsage}. }
\label{tab:sci}
\end{table*}

\paragraph{Retrospection Stage}
We collect trajectories by letting the IL-based LLM interact with the ScienceWorld environment. We then break down each trajectory into steps in the form of \textit{(task description, current state, action, next state)}. More details regarding the collected trajectories and the proportion of the positive trajectories are provided in Table \ref{tab:all-training}. 

For ScienceWorld, the Q-network IQL consists of 1 embedding layer, 5 GRU blocks, and 2 linear layers. The input for the Q-function includes task description, current state, and action. The state is divided into freelook and inventory, the two specific states in ScienceWorld. All these 5 parts (task, current state, action, freelook, and inventory) are passed through separate GPUs. We then concatenate the output of 5 GRU blocks before being fed into the last 2 linear layers.
The $V$-network of IQL is similar to $Q$-network but doesn't need to input action. As a result, the input for $V$ network does not include the action part.

We set the size of the embedding layer to be 64, the output layer of GRU and the first linear layer to be 128. We train the IQL in 20 epochs with a batch size of 128. Due to the light parameter of GRU, the training process is around 2 hours, which is much less than 20h for the warm-up stage. The details of training parameters and training costs are listed in Table \ref{tab:train time IQL} and Table \ref{tab:train paras}.
\paragraph{DRRN}
In ScienceWorld, we separately train one online RL agent (DRRN) for each task of 30 tasks. For each task, the training step for DRRN is 10000, with a learning rate of $1e-4$. The Q-network in DRRN is GRU+MLP with the embedding size of 128 and the hidden size of 128, which is consistent with ScienceWorld paper \cite{wang2022scienceworld}.
 
\paragraph{Extra Results}
Tabel \ref{tab:sci} shows the detailed results for all 30 subtasks of ScienceWorld using different methods. As we can see, Retrospex achieves the highest score in most of the subtasks.

\begin{table*}
\centering
\begin{tabular*}{0.7\linewidth}{ccc}
\toprule
Environment&Warm-up training cost&Retrospection training cost \\ \midrule
SciWorld&4 32G V100s/ 20h&15 rounds/ 1.5h      \\
Webshop&\multirow{2}{*}{4 32G V100s/ 20h}&20 rounds/ <0.5h \\
ALFWorld&&20 rounds/ <0.5h \\
\bottomrule
\end{tabular*}
\caption{Training cost of warm-up stage and retrospection stage on 3 tasks.}
\label{tab:train time IQL}
\end{table*} 

\subsection{Webshop Experimental Details}

\begin{table*}
\centering
\begin{tabular*}{0.75\linewidth}{ccc}
\toprule
Environment&LLM training parameters&IQL training parameters \\ \midrule
SciWorld&Flan-T5-Large(770M)&GRU(2.7M)      \\
Webshop&Llama3-8B-Instruct (Lora,1.5B)&GRU(2.2M)\\
Alfworld&Llama3-8B-Instruct (Lora,1.5B)&GRU(2.2M)\\
\bottomrule
\end{tabular*}
\caption{Training parameters on 3 tasks. The RL critic has a very small size of parameters, thus costing little additional time when doing training and inference.}
\label{tab:train paras}
\end{table*}

\paragraph{Warm-up Stage}\label{web-warm}

Following the approach in \cite{zeng2023agenttuning}, we construct our training data by combining the AgentInstruct and ShareGPT datasets. The inclusion of ShareGPT helps prevent catastrophic forgetting, which could cause LLMs to lose their general capabilities. We refactor the AgentInstruct dataset so that each turn becomes an individual sample. For ShareGPT, we extract samples in a 20:80 ratio relative to AgentInstruct. This results in a training dataset of 13,000 samples from AgentInstruct and 52,000 samples from ShareGPT.

We select the LLaMA3-8B-Instruct model as our backbone model. The training objective follows the approach outlined in \cite{zeng2023agenttuning}. We, however, employ LoRA for fine-tuning, with rank and alpha set to 32 and 64. During fine-tuning, we compute the loss based on the model’s outputs using SFTTrainer\footnote{https://huggingface.co/docs/trl/main/en/sft\_trainer}. We utilize a learning rate of 1e-4 and train for 2 epochs with a batch size of 2. To ensure an efficient training, we leverage DeepSpeed Zero-3 \cite{rajbhandari2021zeroinfinity} for parallel training across four V100 GPUs.

\begin{figure}
    \centering
    \includegraphics[width=0.95\linewidth]{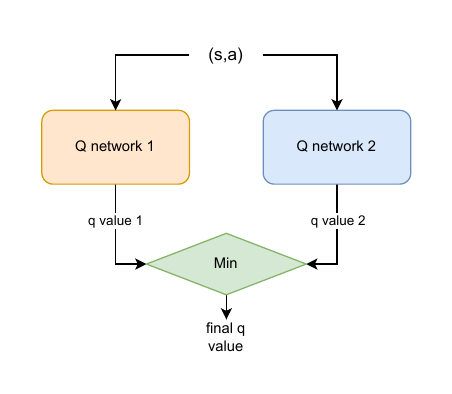}
    \caption{The structure of Twin-Q.}
    \label{twinq}
\end{figure} 
 
\paragraph{Retrospection Stage}
We collect trajectories on Webshop and perform preprocessing similar to that in ScienceWorld. More details are provided in Table \ref{tab:all-training}. For Webshop, we treat the state as a whole part and use one GRU block for it, which is different from ScienceWorld. The Q-network we use is Twin Q (Clipped Double Q-learning) \cite{fujimoto2018addressing}, which uses 2 networks with the same structure and the last $Q$ value is the minimum of these two networks. The structure of Twin Q is shown in Figure \ref{twinq}. The number of steps is comparatively small on Webshop, thus using Twin-Q can make the Q-network more stable. The other part of IQL is the same as that in ScienceWorld.

We set the embedding size of the embedding layer to be 64, the output layer of GRU and the first linear layer to be 128. We train the IQL in 20 epochs with batch size 128. Due to the light parameter of GRU, the training process is around 2 hours. The details of training parameters and cost of warm-up and retrospection stage are listed in Table \ref{tab:train time IQL} and Table \ref{tab:train paras}.


\begin{table}[]
\centering
\begin{tabular}{ccccc}
\toprule
Samples & 500  & 1000  & 1500 & 2000  \\ \midrule
AR      & 76.4 & 76.7  & \textbf{77.7} & 77.3 \\
SR      & 0.475 & 0.475  & \textbf{0.505} & 0.485 \\  \bottomrule
\end{tabular}
\caption{Results of Retrospex in Webshop (AgentLM test set) with IQL trained in different number of collected samples.}
\label{tab:webshoplen}
\end{table}
\paragraph{Analysis on Collected Samples}
Table \ref{tab:webshoplen} shows the impact of the number of samples collected on RL training and the final performance on the Webshop environment. In order to discover the impact of the number of samples collected, we trained different IQLs with a number of samples collected from the Webshop environment ranging from 500 to 2000. We find that using 1500 samples can obtain better performance which means using 2000 samples may have resulted in the problem of overfitting. Due to this consideration, we believe that the training of lightweight IQL needs to be further investigated to achieve a balance between overfitting and underfitting.

\subsection{ALFWorld Experimental Details}
\paragraph{Warm-up Stage}
We train only one model for both ALWorld and Webshop. For detail training information, please refer to section \ref{web-warm}.

\paragraph{Retrospection Stage}
We collect trajectories on ALFWorld and perform preprocessing in the same with ScienceWorld and Webshop. Details are also provided in Table \ref{tab:all-training}. For ALFWorld, the training of IQL is identical to Webshop, where we use Twin-Q for Q-network.



\subsection{More Comparison with $A^3T$}

\begin{table}[]
\centering
\scalebox{0.95}{
\begin{tabular}{ccc}
\toprule
Method & LLM  & SR  \\ \midrule
$A^3T$ (round=0)                    & Mistral-7B-Instruct                    &86.0 \\
$A^3T$ (round=1)                    & Mistral-7B-Instruct                    &94.0 \\
\textbf{$A^3T$ (round=2)}                  & \textbf{Mistral-7B-Instruct}             &\textbf{96.0} \\
$A^3T$ (round=3)                    & Mistral-7B-Instruct                    &95.0 \\
FT-LLaMA3               & LLaMA3-8B-Instruct                 &83.5\\ \midrule
Retrospex               & LLaMA3-8B-Instruct          &87.0 \\ \bottomrule
\end{tabular}}
\caption{Overall results on ALFWorld. The result of the $A^3T$ is from paper \cite{a3t}.}
\label{tab:alfa3t}
\end{table}

\begin{table}[]
\centering
\begin{tabular}{cccc}
\toprule
Method & LLM & AS   \\ \midrule
$A^3T$ (round=0)                    & Mistral 7B                    &72.0    \\
$A^3T$ (round=1)                    & Mistral 7B                    & 73.5  \\
$A^3T$ (round=2)                    & Mistral 7B                    & 72.3   \\
$A^3T$ (round=3)                  & Mistral 7B             & 72.9  \\ 
\midrule
\textbf{Retrospex}  & \textbf{LLaMA3 8B} & \textbf{77.2}   \\    \bottomrule
\end{tabular}
\caption{Overall results on Webshop, Agentboard Testset. The result of the $A^3T$ is from $A^3T$ paper \cite{a3t}.}
\label{tab:all-web-a3t}
\end{table}

We conduct a detailed comparison with $A^3T$ on Webshop and ALFWorld. The results are shown in Table \ref{tab:alfa3t} and \ref{tab:all-web-a3t}. In both environments, \modelname{} outperforms the result of $A^3T$ at $round=0$. Because $A^3T$ continues to increase the training trajectories and trains the LLM at each round, it works better in $round=1,2,3$ in the ALFWorld environment. However, compared to the expensive training investment of $A^3T$, \modelname{} still has advantages.
On Webshop, \modelname{} outperformed all $A^3T$ rounds in 251 test cases of AgentBoard.



\subsection{More Analysis}

\begin{table}
\centering
\scalebox{0.95}{
\begin{tabular}{p{2cm}lcc} \toprule
     \textbf{Env} & \textbf{Methods} & \textbf{AS} & \textbf{SR}  \\ \midrule
     \rowcolor{my_gray}SciWorld & Retrospex & \textbf{55.98} & \textbf{36.0} \\
     &  w/o retrospection & 48.80 & 27.0 \\
     &  w/o act mapping & 55.25 & 34.3 \\ \midrule
     \rowcolor{my_gray} ALFWorld & Retrospex & - & \textbf{87.0} \\
     & w/o retrospection & - & 83.5 \\
     & w/o act mapping & - & 85.1 \\ \midrule
\end{tabular}}
\caption{Results of our ablation study.} 
\label{tab:ablation-study}
\end{table}

Table \ref{tab:ablation-study} shows the role of action mapping in \modelname{} in ScienceWorld and ALFWorld. Here, \modelname{} (w/o retrospection) correspond to IL-T5 in ScienceWorld and \illama{} in ALFWorld, which we put here for cross-reference. As expected, removing action mapping leads to a performance decline in \modelname{}, consistent with previous findings \cite{brohan2023can}. However, the drop is relatively modest, suggesting that action mapping is not the only key factor for \modelname{} in these environments. One possible explanation is that imitation learning during the warm-up phase helps the LLM partially adapt to the target environment, reducing the occurrence of invalid actions.

\end{document}